\documentclass[runningheads]{llncs}

\usepackage{graphicx}
\usepackage{amsmath}
\usepackage{amsfonts}
\usepackage{xcolor}
\usepackage[linesnumbered,algoruled,boxed,lined]{algorithm2e}

\SetCommentSty{mycommfont}
\usepackage[colorlinks=True, linkcolor=red, anchorcolor=blue, citecolor=blue]{hyperref}
\usepackage{svg}
\usepackage{lmodern}
\usepackage{subfigure}
\usepackage{multirow}
\usepackage{tikz}
\usepackage{pgfplots}
\pgfplotsset{compat=1.18}
\graphicspath{{images/}}

\SetKwInOut{Parameter}{Parameters}
\SetKwInOut{INIT}{Initialization}
\let\oldnl\nl
\newcommand{\nonl}{\renewcommand{\nl}{\let\nl\oldnl}}

\begin{document}
\title{Approximation of a Pareto Set Segment Using a Linear Model with Sharing Variables}
\titlerunning{Approximation of a Pareto Set Segment}
%
\author{Ping Guo\inst{1}\orcidID{0000-0001-5412-914X},
    Qingfu Zhang\inst{1}\orcidID{0000-0003-0786-0671},\and Xi Lin  \inst{1}\orcidID{0000-0001-5298-6893}}
\authorrunning{Ping Guo and Qingfu Zhang}
%
\institute{Department of Computer Science, City University of Hong Kong\\
    Hong Kong, China\\
    \email{pingguo5-c@my.cityu.edu.hk}, \email{qingfu.zhang@cityu.edu.hk} and \email{xi.lin@my.cityu.edu.hk}}
\maketitle              
\begin{abstract}

    In many real-world applications, the Pareto Set (PS) of a continuous multiobjective optimization problem can be a piecewise continuous manifold. A decision maker may want to find a solution set that approximates a small part of the PS and requires the solutions in this set share some similarities. This paper makes a first attempt to address this issue. We first develop a performance metric that considers both optimality and variable sharing. Then we design an algorithm for finding the model that minimizes the metric to meet the user's requirements. Experimental results illustrate that we can obtain a linear model that approximates the mapping from the preference vectors to solutions in a local area well.
\end{abstract}

\section{Introduction}\label{sec:intro}

This paper considers the following continuous multiobjective optimization problem (MOP):
\begin{equation}\label{eq:MOP}
    \begin{aligned}
         & \mbox{minimize}   &  & F(x)  = (f_1(x),\ldots, f_m(x)), \\
         & \mbox{subject to} &  & x     \in\Omega,
    \end{aligned}
\end{equation}
where $x$ is the decision variable, $\Omega \subseteq R^n$ is the decision space, $F:\Omega \to R^m$ contains $m$ continuous objective functions $f_1(x), \ldots, f_m(x)$, and $R^m$ is the objective space. Very often, the objectives in MOP~(\ref{eq:MOP}) conflict with each other, and no single solution can optimize them  simultaneously~\cite{miettinen2012nonlinear}. \textit{Pareto optimality} is used to define the best trade-off candidate solutions. The set of all the Pareto optimal solutions is called the \textit{Pareto Set} (PS). Its image in the objective space is called the \textit{Pareto Front} (PF).

Aggregation is an important technique for solving MOPs~\cite{zhang2007moea}. Aggregation methods transform (\ref{eq:MOP}) into some single objective optimization problems. For a preference $\lambda$, an aggregation method aggregates all the $f_i$'s into a scalar objective function, optimizes it and generates a Pareto optimal solution $x(\lambda)$ for the preference vector $\lambda$. Under some conditions, an aggregation method can find all the Pareto optimal solutions. In other words, the PS can be modeled by a function $x=x(\lambda)$. Moreover, under regularity conditions, it is piecewise continuous~\cite{zhang2008rm}.

Given a preference vector $\lambda^0$, a decision maker may be interested only in Pareto optimal solutions around $x(\lambda^0)$. It is reasonable to assume that $x(\lambda)$ is linear around a small neighborhood of $\lambda^0$. Let $x^1=(x^1_1,\ldots,x^1_n)$ ,$x^2=(x^2_1,\ldots,x^2_n) \in R^n$ be two candidate solutions, if $x^1_i=x^2_i$, we say that $x^1$ and $x^2$ share variable $x_i$. In many real-life applications, when the preference changes, it is required to have an approximate Pareto optimal solution for the new preference with as many components the same as the current Pareto optimal solution. This requirement can be essential for reusing existing designs and reducing costs. In engineering design, shared components can support module design~\cite{deb2006innovization} and significantly reduce manufacturing costs. Deb et al.\cite{deb2006innovization} advocate conducting data mining among the obtained Pareto optimal solutions to find useful patterns. To date, no research has been conducted on the integration of shared component requirements into the optimization process.

This paper makes a first attempt to address the issue of shared components. We model it as a problem to use a linear model to approximate a PS segment under the constraint of variable sharing. Much effort has been made to model the Pareto set using a math function~\cite{deb2014integrated,eichfelder2009adaptive,zhou2019pareto}. However, all these existing works aim at approximating the actual Pareto set. Our approach considers the quality of solutions beyond Pareto optimality. We trade Pareto optimality for shared component requirements.
Our major contributions can be summarized as follows:

\begin{itemize}
    \item We study the optimality of a solution set under some shared component constraints instead of Pareto optimality.
    \item We incorporate the user's preference and the requirement on shared components to define a performance metric.
    \item We adopt the framework of MOEA/D to develop an algorithm for finding the model that optimizes the performance metric. This model can generate infinite solutions that satisfy the user's requirements.
\end{itemize}



The rest of the paper is organized as follows: We propose the original version of our performance metric and modification considering variable sharing in Section~\ref{sec:perform}. Then we present the form of the linear model and the connection between variable sharing and the sparsity of the model in Section~\ref{sec:sparse}. We give out the framework of our algorithm and implementation details in Section~\ref{sec:method}. In Section~\ref{sec:exp}, we conduct experiments to validate our algorithm. The last section summarizes the paper and list possible future work directions.

\section{Performance Metric for Local Models}\label{sec:perform}
In this section, we introduce our performance metric that considers both optimality and variable sharing of solutions. We first define a preference vector distribution based on the user-provided preference vector. We then use the expected aggregation value of solutions output by a model as the first part of our metric. Finally, we implicitly define the second part of the metric with regards to variable sharing. Different implementations are possible for the second part. In the next section, we present our implementation using a linear model.

\subsection{Local Approximation Metric}

Consider MOP~(\ref{eq:MOP}). Given:
\vspace{-0.05in}
\begin{itemize}
    \item $\lambda^0$: a preference vector, from the $(m-1)$-D probability simplex;
    \item $N(\lambda^0)$: a neighborhood set of $\lambda^0$; 
    \item $\mathcal{P}$: a probability distribution defined on $N(\lambda^0)$.
\end{itemize}
\vspace{-0.05in}
The neighborhood set can have different structures. In general, the neighborhood of $\lambda$ defines a set of preference vectors that are close to $\lambda$ in Euclidean space. The distribution $\mathcal{P}$ enables us to sample preference vectors from the neighborhood set. In this paper, we use a multi-variate normal distribution as the sampling distribution $\mathcal{P}$. This distribution puts more emphasis on the area that near the user's target solutions.

For any preference vector $\lambda \sim \mathcal{P}$, we can define a sub-problem using aggregation functions. In this paper, we use Chebyshev aggregation. Our metric can be generalized to other aggregation functions. The Chebyshev aggregation value of solution $x$ with preference vector $\lambda$ is by:
\begin{equation}
    \vspace{-0.05in}
    g(x,\lambda) = \max_i \lambda_i|f_i(x) - z_i|,
    \vspace{-0.05in}
\end{equation}
where $z_i$ is the Utopian value for the $i$-th objective, and $\lambda_i$ is the $i$-th component of the preference vector $\lambda$. The associated solution $x^*(\lambda)$ to the above sub-problem is as follows:
\begin{equation}
    \vspace{-0.05in}
    x^*(\lambda) = \arg\min_xg(x,\lambda).
\end{equation}
We can denote the above mapping from the preference vector $\lambda$ to the associated optimal solution as  $x(\lambda)$. Now, we use a model $h_\theta(\lambda)$ parameterized by $\theta$ to predict the associated solution to the sub-problem defined with $\lambda$. Since the solution $x^*(\lambda)$ minimizes the aggregation value $g(x, \lambda)$, we can use the expected aggregation value of the model output to evaluate its optimality. For each preference vector $\lambda \sim \mathcal{P}$, the aggregation value computed using the model output is $g(h_\theta(\lambda), \lambda)$. Our metric $\mathcal{M}$ is as follows:
\vspace{-0.05in}
\begin{equation}\label{eq:pre_metric}
    \mathcal{M}(h_\theta) = \mathbb{E}_{\lambda\sim\mathcal{P}}[g(h_\theta(\lambda), \lambda)].
    \vspace{-0.05in}
\end{equation}
The above metric can be directly used to learn the Pareto set for different real-world applications, such as multi-task learning~\cite{lin2020controllable}, neural multiobjective combinatorial optimization~\cite{lin2021pareto}, and multiobjective Bayesian optimization~\cite{lin2022pareto}. In this work, we extend it to include the variable sharing constraint.

\subsection{Shared Variable Metric}

The above metric only considers the optimality of solutions under sub-problems defined with different preference vectors. We need to add extra terms to evaluate the model's performance according to special requirements from the user. In this paper, we consider variable sharing due to its importance in engineering design.

We denote the function that measures the degree of variable sharing of the model $h_\theta$ as $\mathcal{I}(h_\theta)$, called variable sharing degree (VSD). Without loss of generality, we assume a lower value of $\mathcal{I}$ indicates a higher degree of variable sharing. The explicit form of the VSD can be various, we will connect it with the sparsity of the model in the next section.

Now our goal is to find the model that minimizes both $\mathcal{M}$ and $\mathcal{I}$. We synthese these two metrics and reload the notion $\mathcal{M}$ to represent our final performance metric. The final version of the metric is as follows:
\begin{equation}\label{eq:metric}
    \mathcal{M}(h_\theta) = \mathbb{E}_{\lambda\sim\mathcal{P}}[g(h_\theta(\lambda), \lambda)] + \gamma \mathcal{I}(h_\theta),
\end{equation}
where $\gamma$ is the parameter that weighs the importance of the VSD. Larger values of $\gamma$ lead to models that trade optimality for variable sharing. Our goal is to build a linear model that can approximate the local area of the PS and produce solutions with as many variables taking the same value across the solutions.

\section{Linear Sparse Representation of the Local PS}\label{sec:sparse}
In this section, we first give the analytical form of our linear model. Then, we discuss the sparsity of the parameters and use it as an implementation of the VSD.

\begin{figure}[t]
    \centering
    \fontsize{6pt}{7pt}\selectfont
    \includesvg[width=0.95\textwidth]{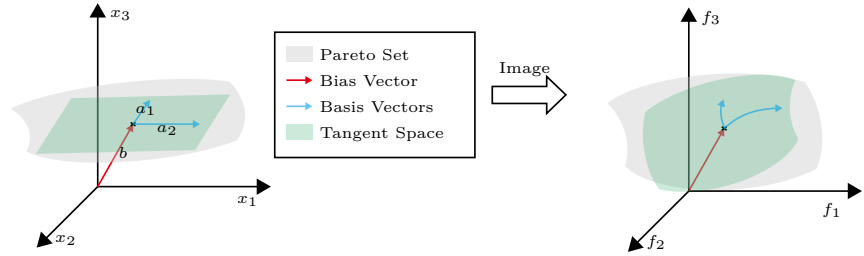}
    \caption{\textbf{Linear Model:} Column vectors (e.g. $a_1, a_2$) in $A$ can be regarded as basis vectors that span the subspace of a hyperplane. This hyperplane can be used to approximate the local PS segment since we expect it to be linear under mild conditions. \label{fig:tangent_space}}
\end{figure}
\vspace{-0.1in}

\subsection{Linear Model}

Since the preference vectors are from a probability simplex, the sum of all the elements is equal to 1. So we only use the first $(m-1)$ elements of the preference vectors $\lambda$, denoted as $\lambda_{1:m-1}$ as the input. Our model designed using \textit{first-order} approximation is as follows:
\vspace{-0.1in}
\begin{equation}\label{eq:linear_model}
    h_\theta(\lambda) = A(\lambda_{1:m-1}-\lambda_{1:m-1}^0) + b,
\end{equation}
where $A\in R^{n\times(m-1)}$ and $b\in R^n$ are the parameters of the model. $\theta=(A, b)$ is still used to represent all the parameters of the model for brevity.

The interpretation of the column vectors of $A$ can be a set of basis vectors of the tangent space at point $x^0$. The bias vector $b$ is used to represent the associated solution $x^0$ with the preference vector $w^0$. We give an illustrative example of the linear model in Fig.~\ref{fig:tangent_space}.

The performance of the above model can be evaluated using metric~(\ref{eq:metric}). However, the form of VSD is not explicitly defined. In the following paragraphs, we give our implementation of VSD using sparsity of the model.

\subsection{Variable Sharing and Sparsity of the model}

We notice that the linear approximation is actually a set of linear combinations of the column vectors in $A$ and the bias vector $b$. Each non-zero row in $A$ contributes to one dimension of output solutions the model. More ``empty" rows in $A$ lead to more shared variables in the solutions output by $h_\theta(\lambda)$. Therefore, we use the row sparsity of $A$ as an implementation of the VSD. Specifically, we use the $(2,1)$-norm of matrix $A$ as the function to measure the degree of row sparsity. The $(2,1)$-norm of matrix $A$ is defined as:
\begin{equation}
    \|A\|_{2,1}=\sum_{i=1}^n\|a_i\|_2,
\end{equation}
where $a_i$ is the $i$-th row of matrix $A$.

\section{Method and Algorithm}\label{sec:method}
\subsection{An Alternative Problem}
Using the linear model defined in the previous section, the optimization problem becomes:
\begin{equation}
    \min_{A, b}\mathcal{M}(h_\theta) = \mathbb{E}_{\lambda\sim\mathcal{P}}[g(h_\theta(\lambda), \lambda)] + \gamma \|A\|_{2,1}.
\end{equation}

If the explicit form of $F$ is known, we can derive the form of $g$ accordingly and apply a gradient descent algorithm to solve it. However, for real-world problems, the gradient information is often unavailable.

Our solution is to maintain a dynamic dataset of preference vector-solution pairs during our algorithm. We update this dataset in each iteration, assuming it converges to the local PS progressively. Then, the data from it can be regarded as noisy samples from the true PS. We replace the first term with the mean squared loss (MSE) between the samples and the output of the model.

Suppose we have a dataset $\left\{(\lambda^1,x^1),\ldots,(\lambda^N,x^N)\right\}$ at any iteration of our algorithm. We want to find a linear model to fit this dataset. More specifically,
our goal is to minimize the following loss function:
\begin{equation}
    \frac{1}{N}\sum_{i=1}^N\|x^i-(A(\lambda_{1:m-1}^i-\lambda_{1:m-1}^0) + b)\|^2+\gamma\|A\|_{2,1},
\end{equation}
where $\lambda^0$ is the user-given preference vector.

The above problem is convex and easy to solve. In fact, we can minimize the above loss function by solving a series of simple regularized least square regression problems for each dimension of the data.

\subsection{Algorithm Framework}
\subsubsection{Initialization of the Preference Set} In the previous section, we described how a dataset of preference vector-solution pairs is used to build the linear model. Here we illustrate the process of initializing the preference set using the user-given preference vector $\lambda^0$:
\begin{itemize}
    \item Use a normal distribution $\mathcal{N}(0, \sigma^2 I)$ to sample noise vectors.
    \item Sample $N$ noise vectors from $\mathcal{N}(0, \sigma^2 I)$, add $\lambda^0$ to them to generate $N$ vectors.
    \item Project these vectors onto the probability simplex to normalize the disturbed vectors.
\end{itemize}

\begin{figure}[htbp]
    \centering
    \fontsize{6pt}{7pt}\selectfont
    \includesvg[width=0.65\textwidth]{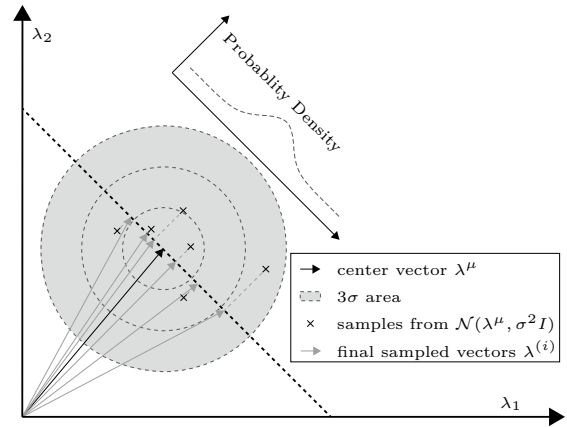}
    \caption{\textbf{Preference Vector Generation:} This process is equivalent to sampling from $\mathcal{N}(\lambda^0, \sigma I)$ and projecting them on the simplex.\label{fig:sample_weight}}
\end{figure}

Through the above generation process, we obtain a preference set of $N$ preference vectors. This process can also be viewed as sampling from the neighborhood of the preference vector on the simplex, as shown in Fig.~\ref{fig:sample_weight}.

\vspace{-0.5cm}

\subsubsection{Main Algorithm} In general, we adopt the framework of MOEA/D~\cite{zhang2007moea} to design our algorithm. Details of our MOEA/D with local linear approximations, called MOEA/D-LLA (Local Linear Approximation), are shown in Algorithm~\ref{alg:our_alg}. It takes the obtained preference vector set as its input. We first initialize a set of sub-problems using the preference set under Chebyshev aggregation and assign the value of reference point as done in MOEA/D. Then our algorithm maintains:

\begin{itemize}
    \item a population $X$ of size $N$, where the $i$-th individual is used to solve the sub-problem using $\lambda^i$,
    \item a set of decomposed value $\left\{g(x^i,\lambda^i)|1\leq i \leq N\right\}$,

    \item and a reference point $z^*=(z^*_1,\ldots, z^*_m)^T$.
\end{itemize}
We generally push the population towards the PS by using genetic operators and our model to generate new solutions. We train our model by solving a regression task on the dataset of preference vector-solution pairs.
\begin{algorithm}[htbp]
    \caption{MOEA/D-LLA\label{alg:our_alg}} 
    \KwIn{preference vector set $W = \left\{\lambda^1,\ldots, \lambda^N\right\}$}
    \Parameter{regularization parameter $\gamma$, optimization step $o$}
    \KwOut{matrix $A$, bias vector $b$, solution set $X$}
    Initialize matrix $A$, $b$, a population $X$\\
    \While{not terminated}{
    \tcp{Optimization Step}
    \label{lst:line:blah1}Using MOEA/D to optimize $g(x, \lambda^1), \ldots, g(x, \lambda^N)$ to obtain $(\lambda^1, x^1), \ldots, (\lambda^N, x^N)$; \\
    \tcp{Regression Step}
    \For{$i = 1 \to o$}{
    $A^*, b^* = \arg\min_{A, b}{\frac{1}{N}\sum_{i=1}^{N}\left\|x^i - (A(\lambda^i_{1:m-1}-\lambda^0_{1:m-1})+b)\right\|^2_2+ \gamma \|A\|_{2,1}}$
    }
    \tcp{Update Population Using Linear Model}
    \label{lst:line:blah2}Generate new solutions using Algorithm~\ref{alg:new_sol} and use them to update population $(\lambda^1, x^1), \ldots, (\lambda^N, x^N)$.\\
    }
\end{algorithm}
\vspace{-0.5cm}

\subsubsection{Sampling New Solutions} In our algorithm, we use a hybrid strategy to generate new solutions. In Line~\ref{lst:line:blah1}, we use genetic operators to sample new solutions. In Line~\ref{lst:line:blah2}, we sample solutions from the linear model as in \cite{zhou2019pareto}. The sampling method is given in Algorithm~\ref{alg:new_sol}.

\begin{algorithm}[htbp]
    \caption{Sampling New Solutions\label{alg:new_sol}}
    Sample $N$ noise vectors from  $\mathcal{N}(0, \sigma_{noise}^2 I)$;\\
    Add noise vectors on the preference set to obtain the noised set $\left\{\tilde{\lambda}^1, \ldots, \tilde{\lambda}^N\right\}$;\\
    Generate $N$ solutions using
    \begin{equation*}
        x = A\tilde{\lambda}_{1:m-1} + b
    \end{equation*}
\end{algorithm}
\vspace{-0.5cm}

\section{Experimental Results}\label{sec:exp}
In this section, we study the optimality and the variable sharing aspect of our algorithm. Since standard test instances like ZDT~\cite{zitzler2000comparison} and DTLZ~\cite{deb2005scalable} naturally have shared variables in their Pareto set, we first design a problem with no shared variables and test our algorithm on it. Then we study the trade-off between optimality and variable sharing for this problem. Here, we use R-metric from \cite{li2017r} to incorporate the user-given preference vector into the evaluation of the solutions. The parameter $\delta$ for R-metric is set to be $6\sigma$ to target our preferred area.

\subsection{Parameter Setting}
All the experimental results are obtained from 10 independent runs of the algorithm. We add extra function evaluations in MOEA/D-DE as a compensation for not generating solutions from the model to ensure a fair comparison. The parameters of our algorithm are set as follows:
\begin{itemize}
    \item The population size: It is 100 for all instances.
    \item The variance $\sigma^2$ of the sampling distribution $\mathcal{P}$: It is 0.02 for all instances.
    \item The variance $\sigma_{noise}^2$ for sampling new solutions: It is 0.05 for all instances.
    \item The maximum number of generations: It is 300 for all instances.
\end{itemize}

\subsection{Performance on None-shared Problem}

\begin{figure}[htbp]
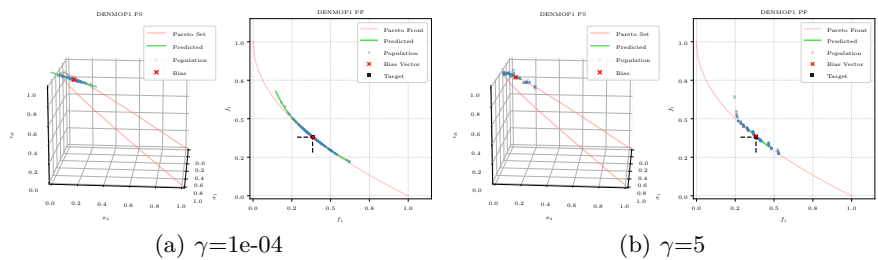

    \centering
    \subfigure[$\gamma$=1e-04]{
        \centering
        \fontsize{2pt}{3pt}\selectfont
        \includesvg[width=0.23\textwidth]{DENMOP1_n10m2_ps_pred_300_1e-4.svg}
        \includesvg[width=0.23\textwidth]{DENMOP1_n10m2_pf_pred_300_1e-4.svg}
    }
    \subfigure[$\gamma$=5]{
        \centering
        \fontsize{2pt}{3pt}\selectfont
        \includesvg[width=0.23\textwidth]{DENMOP1_n10m2_ps_pred_300_5.svg}
        \includesvg[width=0.23\textwidth]{DENMOP1_n10m2_pf_pred_300_5.svg}
    }
    \caption{\textbf{Linear Approximation for Local PS:} An illustration of the population and the predictions of the model for a nonlinear Pareto set under different $\gamma$ in both decision space and objective space.\label{fig:res_both}}

\end{figure}

To evaluate the effectiveness of our algorithm in finding solutions with shared variables, we create a new test instance, MOZDT1, by modifying ZDT1 by replacing $g(x)$. The form of our $g(x)$ is as follows:

\begin{equation}
    \begin{aligned}
        l(x) & = ((1 - 2x_1)^2 -x_2)^2 + (x_3 + x_2 - 1)^2        \\
        g(x) & = 1 + \frac{9}{n-3}\sum_{i=4}^n |x_i - x_1| + l(x)
    \end{aligned}
\end{equation}

The Pareto set for this problem is defined as:
\begin{equation}
    \begin{aligned}
        0\leq                   x_1 \leq 1, x_2=(1 -2x_1)^2, x_3 = 1-  (1 -2x_1)^2, \text{and }x_i=x_1(i=4,\ldots, n)
    \end{aligned}
\end{equation}

The Pareto optimal solutions in this set have no shared variables. Therefore, we cannot guarantee the optimality of the solutions if we desire more shared variables in them. As $\gamma$ controls the importance of VSD in (\ref{eq:metric}), we expect our model's output have more shared variables as $\gamma$ increases.

\begin{figure}[htbp]
    \centering
    \subfigure[\label{fig:res_den_1}]{
        \centering
\begin{tikzpicture}[scale=0.5]

  \definecolor{darkgray176}{RGB}{176,176,176}
  \definecolor{darkorange25512714}{RGB}{255,127,14}
  \definecolor{lightgray204}{RGB}{204,204,204}
  \definecolor{steelblue31119180}{RGB}{31,119,180}

  \begin{axis}[
      legend cell align={left},
      legend style={
          fill opacity=0.8,
          draw opacity=1,
          text opacity=1,
          at={(0.03,0.97)},
          anchor=north west,
          draw=lightgray204
        },
      tick align=outside,
      tick pos=left,
      title={R-IGD},
      x grid style={darkgray176},
      xlabel={\(\displaystyle \gamma\)},
      xmajorgrids,
      xmin=-4.2349485002168, xmax=0.93391850455282,
      xtick style={color=black},
      xtick={-4,-3,-1.30102999566398,0,0.698970004336019},
      xticklabels={1e-04,1e-03,5e-02,1e+00,5e+00},
      y grid style={darkgray176},
      ylabel={R-IGD},
      ymajorgrids,
      ymin=0.0860836308706898, ymax=0.302695178030354,
      ytick style={color=black}
    ]
    \path [fill=steelblue31119180, fill opacity=0.2]
    (axis cs:-4,0.120077000703284)
    --(axis cs:-4,0.103183642796716)
    --(axis cs:-3,0.102899318071116)
    --(axis cs:-1.30102999566398,0.107833959805449)
    --(axis cs:0,0.146458011796783)
    --(axis cs:0.698970004336019,0.173192196047475)
    --(axis cs:0.698970004336019,0.190759374952525)
    --(axis cs:0.698970004336019,0.190759374952525)
    --(axis cs:0,0.189046762703217)
    --(axis cs:-1.30102999566398,0.131306252694551)
    --(axis cs:-3,0.126083403428884)
    --(axis cs:-4,0.120077000703284)
    --cycle;

    \path [fill=darkorange25512714, fill opacity=0.2]
    (axis cs:-4,0.112429759212962)
    --(axis cs:-4,0.0959296102870382)
    --(axis cs:-3,0.097670649857858)
    --(axis cs:-1.30102999566398,0.107409460721411)
    --(axis cs:0,0.166921812983846)
    --(axis cs:0.698970004336019,0.268900061385994)
    --(axis cs:0.698970004336019,0.292849198614006)
    --(axis cs:0.698970004336019,0.292849198614006)
    --(axis cs:0,0.214319354016154)
    --(axis cs:-1.30102999566398,0.136793684778589)
    --(axis cs:-3,0.123808705642142)
    --(axis cs:-4,0.112429759212962)
    --cycle;

    \addplot [semithick, steelblue31119180, mark=x, mark size=3, mark options={solid}]
    table {%
        -4 0.11163032175
        -3 0.11449136075
        -1.30102999566398 0.11957010625
        0 0.16775238725
        0.698970004336019 0.1819757855
      };
    \addlegendentry{R-IGD(pop)}
    \addplot [semithick, darkorange25512714, mark=triangle*, mark size=3, mark options={solid}]
    table {%
        -4 0.10417968475
        -3 0.11073967775
        -1.30102999566398 0.12210157275
        0 0.1906205835
        0.698970004336019 0.28087463
      };
    \addlegendentry{R-IGD(pred)}
  \end{axis}

\end{tikzpicture}
    }
    \subfigure[\label{fig:res_den_2}]{
        \centering
        \fontsize{4pt}{5pt}\selectfont
        \includesvg[width=0.5\textwidth]{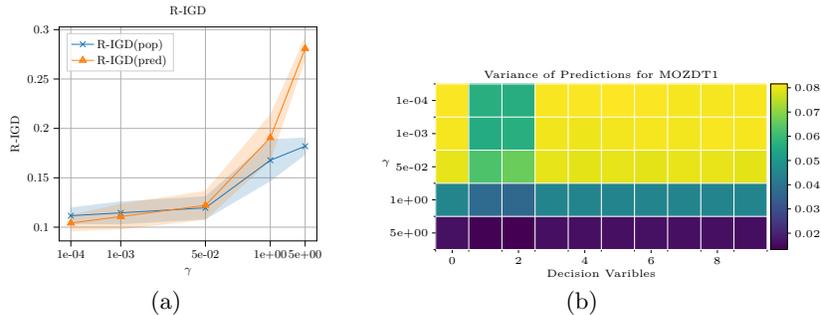}
    }
    \caption{(a) The value of R-IGD calculated on the population from the last iteration and the predictions given by the model. The area within one standard deviation is shaded. (b) The variance of decision variables of the predictions. The model's predictions converge to a small area as $\gamma$ increases.\label{fig:res_den}}
\end{figure}

We run our algorithms under different $\gamma$ values and plot the results in Fig.~\ref{fig:res_both}. Additionally, we show the optimality and the degree of variable sharing in Fig.~\ref{fig:res_den}. In Fig.~\ref{fig:res_den}, we use R-IGD to evaluate the optimality of the solutions and the variances of decision variables to illustrate VSD.

Fig.~\ref{fig:res_den_1} shows that with a small gamma value, our algorithm is able to find a model that can generate solutions of high quality. However, we observe a significant deterioration of optimality when $\gamma$ was increased to 5. Upon examining the predictions associated with $\gamma=5$, we find that their variances reduce to almost zero. This indicates that most solutions output by the model are very similar to each other and can be considered the same solution. In future work, We will further investigate the impact of $\gamma$ on different decision variables.

\subsection{Performance on Standard Test Instances}
The Pareto optimal solutions of standard test instances like ZDT~\cite{zitzler2000comparison} and DTLZ~\cite{deb2005scalable} have special structures in which the majority of their decision variables are shared. Therefore, if $\gamma$ is set correctly, we expect VSD to act as a regularization term and help obtain a model that balances optimality and variable sharing.

We evaluate the quality of the solutions produced by MOEA/D-DE, our algorithm, and the model's predictions using the R-metric. The results are listed in Table~\ref{tab:metric}.
\begin{table}[htbp]
    \caption{R-IGD and R-HV values obtained by MOEA/D-DE and LLA over 10 independent runs. The value of the population and the predictions are both evaluated.\label{tab:metric}}
    \centering
    \begin{tabular}{lcccccc}
        \hline
        \multirow{2}{*}{Problem} & \multicolumn{2}{c}{MOEA/D-DE} & \multicolumn{2}{c}{LLA Pop} & \multicolumn{2}{c}{LLA Pred}                                                    \\\cline{2-7}
                                 & R-IGD                         & R-HV                        & R-IGD                        & R-HV     & R-IGD             & R-HV              \\\hline
        ZDT1                     & 1.27e-01                      & 5.56e-01                    & 1.27e-01                     & 5.56e-01 & \textbf{7.76e-02} & \textbf{6.09e-01} \\
        ZDT2                     & 1.02e-01                      & 2.89e-01                    & 1.02e-01                     & 2.89e-01 & \textbf{6.60e-02} & \textbf{2.95e-01} \\
        ZDT4                     & 1.26e-01                      & 5.55e-01                    & 1.26e-01                     & 5.55e-01 & \textbf{7.78e-02} & \textbf{6.08e-01} \\
        ZDT6                     & 1.26e-01                      & 3.37e-01                    & 1.16e-01                     & 3.42e-01 & \textbf{8.27e-02} & \textbf{3.46e-01} \\
        DTLZ1                    & 1.90e-01                      & 5.09e-01                    & 1.92e-01                     & 5.05e-01 & \textbf{1.84e-01} & \textbf{6.36e-01} \\
        DTLZ2                    & 3.24e-01                      & 1.90e-01                    & 3.24e-01                     & 1.90e-01 & \textbf{2.82e-01} & \textbf{2.30e-01} \\
        DTLZ3                    & 3.23e-01                      & 1.90e-01                    & 3.25e-01                     & 1.90e-01 & \textbf{2.82e-01} & \textbf{2.29e-01} \\
        DTLZ4                    & \textbf{3.97e-01}             & \textbf{1.48e-01}           & 4.22e-01                     & 1.35e-01 & 4.22e-01          & 1.42e-01          \\\hline
    \end{tabular}
\end{table}

The results in Table.~\ref{tab:metric} show that LLA's predictions achieve the best R-IGD and R-HV values in 7 out of 8 standard test instances. Moreover, the populations of LLA outperform those of the original MOEA/D in terms of R-metrics. This superior performance can be attributed to the fact that the problems' Pareto optimal solutions naturally share most decision variables. Therefore, adding variable sharing constraint does not significantly degrade the performance of the LLA algorithm.

\begin{figure}[htbp]
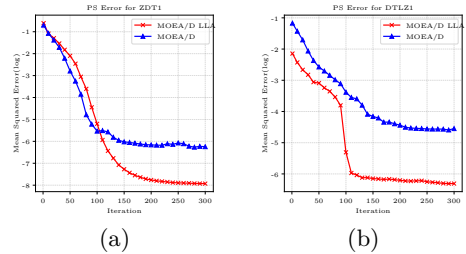

    \centering
    \subfigure[]{
        \centering
        \fontsize{2.5pt}{3pt}\selectfont
        \includesvg[width=0.25\textwidth]{./ZDT1_n30m2_ps_error.svg}
    }
    \subfigure[]{
        \centering
        \fontsize{2.5pt}{3pt}\selectfont
        \includesvg[width=0.25\textwidth]{./DTLZ1_n7m3_ps_error.svg}
    }
    \caption{Mean squared errors between the true optimal solutions and the model's predictions for ZDT1 and DTLZ1. The results are the average value of 10 independent runs and are plotted on logarithmic axis.\label{fig:error}}
\end{figure}

\textbf{Approximation Error} To further evaluate the convergence of our algorithm on these instances, we plot the mean squared error between the output of our model and the true optimal solutions associated with the sub-problems defined by the preference vectors in Fig.~\ref{fig:error}. In these experiments, $\gamma$ is set to be a small value (1e-03).

From Fig.~\ref{fig:error}, we can see that the residual error of MOEA/D-DE can not be further diminished by increasing the number of iterations. Our model's predictions improve quickly and exceed the performance of MOEA/D-DE after around 100 iterations. Furthermore, the quality of the solution for each sub-problem continues to improve with more function evaluations.

\begin{figure}[htbp]
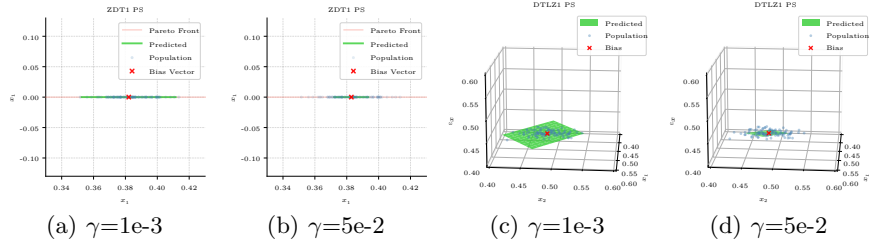

    \centering
    \subfigure[$\gamma$=1e-3\label{fig:ps_share_1}]{
        \centering
        \fontsize{2.5pt}{3pt}\selectfont
        \includesvg[width=0.22\textwidth]{./ZDT1_n30m2_ps_pred_300_0.001.svg}
    }
    \subfigure[$\gamma$=5e-2\label{fig:ps_share_2}]{
        \centering
        \fontsize{2.5pt}{3pt}\selectfont
        \includesvg[width=0.22\textwidth]{./ZDT1_n30m2_ps_pred_300_0.04.svg}
    }
    \subfigure[$\gamma$=1e-3\label{fig:ps_share_3}]{
        \centering
        \fontsize{2.5pt}{3pt}\selectfont
        \includesvg[width=0.22\textwidth]{./DTLZ1_n7m3_ps_pred_300_0.001.svg}
    }
    \subfigure[$\gamma$=5e-2\label{fig:ps_share_4}]{
        \centering
        \fontsize{2.5pt}{3pt}\selectfont
        \includesvg[width=0.22\textwidth]{./DTLZ1_n7m3_ps_pred_300_0.04.svg}
    }
    \caption{The influence of $\gamma$ on the variable sharing degree for ZDT1 and DTLZ1. The shape of the approximated PS shrinks as $\gamma$ becomes bigger.\label{fig:ps_share}}
\end{figure}

\textbf{Variable Sharing} The weighting factor $\gamma$ controls the trade-off between optimality and variable sharing. We plot the population from the last iteration and the hyper-plane that represents the output of the linear model obtained with different values of $\gamma$ in Fig.~\ref{fig:ps_share}.

With larger $\gamma$, the variance of the decision variable $x_2$ becomes smaller in the output of the models. We see the same phenomenon for different test instances in Fig.~\ref{fig:ps_share_3} and Fig.~\ref{fig:ps_share_4}. We can conclude now that our model has a higher degree of variable sharing when we increase $\gamma$. Moreover, these examples illustrate an enhancement of the influence of $\gamma$ for PS in the higher-dimensional space.
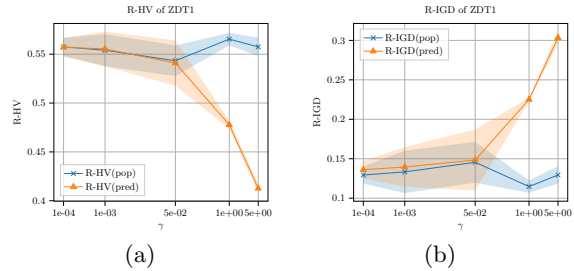
\begin{figure}[htbp]
    \centering
    \subfigure[]{
        \centering
        \resizebox{0.30\textwidth}{!}{
\begin{tikzpicture}

    \definecolor{darkgray176}{RGB}{176,176,176}
    \definecolor{darkorange25512714}{RGB}{255,127,14}
    \definecolor{lightgray204}{RGB}{204,204,204}
    \definecolor{steelblue31119180}{RGB}{31,119,180}

    \begin{axis}[
            legend cell align={left},
            legend style={
                    fill opacity=0.8,
                    draw opacity=1,
                    text opacity=1,
                    at={(0.03,0.03)},
                    anchor=south west,
                    draw=lightgray204
                },
            tick align=outside,
            tick pos=left,
            title={R-HV of ZDT1},
            x grid style={darkgray176},
            xlabel={\(\displaystyle \gamma\)},
            xmajorgrids,
            xmin=-4.2349485002168, xmax=0.93391850455282,
            xtick style={color=black},
            xtick={-4,-3,-1.30102999566398,0,0.698970004336019},
            xticklabels={1e-04,1e-03,5e-02,1e+00,5e+00},
            y grid style={darkgray176},
            ylabel={R-HV},
            ymajorgrids,
            ymin=0.399521013583177, ymax=0.581465805128283,
            ytick style={color=black}
        ]
        \path [fill=steelblue31119180, fill opacity=0.2]
        (axis cs:-4,0.566817397298929)
        --(axis cs:-4,0.548231080201071)
        --(axis cs:-3,0.53760218387339)
        --(axis cs:-1.30102999566398,0.527562885924324)
        --(axis cs:0,0.559029506658535)
        --(axis cs:0.698970004336019,0.548163814446988)
        --(axis cs:0.698970004336019,0.566794694053012)
        --(axis cs:0.698970004336019,0.566794694053012)
        --(axis cs:0,0.571811992341465)
        --(axis cs:-1.30102999566398,0.558989835575676)
        --(axis cs:-3,0.57053036212661)
        --(axis cs:-4,0.566817397298929)
        --cycle;

        \path [fill=darkorange25512714, fill opacity=0.2]
        (axis cs:-4,0.566573269171236)
        --(axis cs:-4,0.547562328828764)
        --(axis cs:-3,0.537565690169222)
        --(axis cs:-1.30102999566398,0.517654704313167)
        --(axis cs:0,0.473662762570326)
        --(axis cs:0.698970004336019,0.407791231380682)
        --(axis cs:0.698970004336019,0.417542177619318)
        --(axis cs:0.698970004336019,0.417542177619318)
        --(axis cs:0,0.481895070429674)
        --(axis cs:-1.30102999566398,0.564180405186833)
        --(axis cs:-3,0.573195587330778)
        --(axis cs:-4,0.566573269171236)
        --cycle;

        \addplot [semithick, steelblue31119180, mark=x, mark size=3, mark options={solid}]
        table {%
                -4 0.55752423875
                -3 0.554066273
                -1.30102999566398 0.54327636075
                0 0.5654207495
                0.698970004336019 0.55747925425
            };
        \addlegendentry{R-HV(pop)}
        \addplot [semithick, darkorange25512714, mark=triangle*, mark size=3, mark options={solid}]
        table {%
                -4 0.557067799
                -3 0.55538063875
                -1.30102999566398 0.54091755475
                0 0.4777789165
                0.698970004336019 0.4126667045
            };
        \addlegendentry{R-HV(pred)}
    \end{axis}

\end{tikzpicture}
        }
    }
    \subfigure[]{
        \centering
        \resizebox{0.30\textwidth}{!}{
\begin{tikzpicture}

  \definecolor{darkgray176}{RGB}{176,176,176}
  \definecolor{darkorange25512714}{RGB}{255,127,14}
  \definecolor{lightgray204}{RGB}{204,204,204}
  \definecolor{steelblue31119180}{RGB}{31,119,180}

  \begin{axis}[
      legend cell align={left},
      legend style={
          fill opacity=0.8,
          draw opacity=1,
          text opacity=1,
          at={(0.03,0.97)},
          anchor=north west,
          draw=lightgray204
        },
      tick align=outside,
      tick pos=left,
      title={R-IGD of ZDT1},
      x grid style={darkgray176},
      xlabel={\(\displaystyle \gamma\)},
      xmajorgrids,
      xmin=-4.2349485002168, xmax=0.93391850455282,
      xtick style={color=black},
      xtick={-4,-3,-1.30102999566398,0,0.698970004336019},
      xticklabels={1e-04,1e-03,5e-02,1e+00,5e+00},
      y grid style={darkgray176},
      ylabel={R-IGD},
      ymajorgrids,
      ymin=0.096190401696069, ymax=0.321456020632703,
      ytick style={color=black}
    ]
    \path [fill=steelblue31119180, fill opacity=0.2]
    (axis cs:-4,0.140313889469709)
    --(axis cs:-4,0.118408852030291)
    --(axis cs:-3,0.106429748011371)
    --(axis cs:-1.30102999566398,0.119627057514388)
    --(axis cs:0,0.106780632246173)
    --(axis cs:0.698970004336019,0.118449246933553)
    --(axis cs:0.698970004336019,0.140485148066447)
    --(axis cs:0.698970004336019,0.140485148066447)
    --(axis cs:0,0.122672046753827)
    --(axis cs:-1.30102999566398,0.171375761485612)
    --(axis cs:-3,0.160194822488629)
    --(axis cs:-4,0.140313889469709)
    --cycle;

    \path [fill=darkorange25512714, fill opacity=0.2]
    (axis cs:-4,0.147281763646476)
    --(axis cs:-4,0.125068685853524)
    --(axis cs:-3,0.114379964034987)
    --(axis cs:-1.30102999566398,0.109659550142393)
    --(axis cs:0,0.222920324301966)
    --(axis cs:0.698970004336019,0.295641185682599)
    --(axis cs:0.698970004336019,0.311216674317401)
    --(axis cs:0.698970004336019,0.311216674317401)
    --(axis cs:0,0.227206650198034)
    --(axis cs:-1.30102999566398,0.187114713857606)
    --(axis cs:-3,0.164703501465013)
    --(axis cs:-4,0.147281763646476)
    --cycle;

    \addplot [semithick, steelblue31119180, mark=x, mark size=3, mark options={solid}]
    table {%
        -4 0.12936137075
        -3 0.13331228525
        -1.30102999566398 0.1455014095
        0 0.1147263395
        0.698970004336019 0.1294671975
      };
    \addlegendentry{R-IGD(pop)}
    \addplot [semithick, darkorange25512714, mark=triangle*, mark size=3, mark options={solid}]
    table {%
        -4 0.13617522475
        -3 0.13954173275
        -1.30102999566398 0.148387132
        0 0.22506348725
        0.698970004336019 0.30342893
      };
    \addlegendentry{R-IGD(pred)}
  \end{axis}

\end{tikzpicture}
        }
    }
    \caption{An illustration of the value of R-metric of the solutions set and the predictions of the linear model of our algorithm. The lines are the average value of 10 independent runs, while the shaded areas reflect the variance of the value.\label{fig:influence}}
\end{figure}

We further investigate the influence of $\gamma$ on optimality by plotting $\gamma$ against the R-metric value of ZDT1 problem in Fig.~\ref{fig:influence}. The R-metrics of the solution set fluctuate with the value of $\gamma$. However, we observe significant deterioration of performance of the linear model when we increase $\gamma$. Although ZDT1's Pareto optimal solutions naturally share the first decision variable, higher degree of variable sharing leads to worse optimality. We can conclude that the cost of sharing lies in the non-shared variables.

\begin{figure}[htbp]
    \centering
    \includesvg[width=0.9\textwidth]{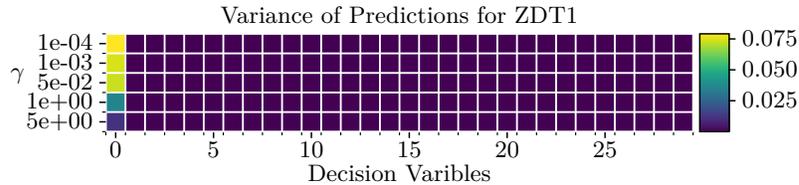}
    \caption{Illustration of the variances of decision variables under different $\gamma$. For ZDT1, most of the decision variables are shared in the Pareto set.\label{fig:heatmap}}
\end{figure}

To better understand the imapct of $\gamma$ on decision variables, we calculate the variance of each decision variable under different $\gamma$ and show them in a heatmap in Fig.~\ref{fig:heatmap}. Smaller variances indicate a higher degree of variable sharing. With a smaller $\gamma$, we can obtain a model that targets the correct shared decision variables without degrading the optimality of the solutions as shown in Fig.~\ref{fig:heatmap}. However, when we increase $\gamma$ to a large value (e.g. 5) to emphasize the importance of variable sharing, we indeed trade the performance (Fig.~\ref{fig:influence}) for variable sharing (Fig.~\ref{fig:heatmap}).

\section{Conclusion}\label{sec:conc}
In this paper, we have studied how to approximate a small part of the PS subject to the variable sharing constraint. We have defined its performance metric as the expectation of the aggregation value under Chebyshev aggregation. Our proposed algorithm can find the optimal linear model that minimizes this performance metric and learns a sparse representation of the local PS. We have conducted  experimental studies on the trade-off between optimality and variable sharing.
In the future, we plan to study the following:
\begin{itemize}
    \item We will consider more test instances where no decision variables are shared in Pareto optimal solutions, further investigate the best trade-off between optimality and variable sharing in different problems settings.
    \item Instead of using regularized least square regression to learn the model parameters, we will explore more efficient and intelligent approaches such as deep learning and reinforcement learning.
\end{itemize}

\subsection*{Acknowledgments}
This work is supported by the General Research Fund (GRF) grant (CityU 11215622) from the Research Grants Council of Hong Kong, China.

%
%
%
\bibliographystyle{splncs04}
\bibliography{emo_paper}

\end{document}